\documentclass[a4paper, 10pt, conference]{ieeeconf}      

\IEEEoverridecommandlockouts
\usepackage{multirow}
\usepackage{cite}
\usepackage{amsmath,amssymb,amsfonts}
\usepackage{graphicx}
\usepackage{textcomp}
\usepackage{tabularx}
\usepackage{amsmath}
\usepackage{dblfloatfix}
\usepackage{siunitx}
\usepackage{booktabs}
\usepackage{longtable}
\usepackage{float}
\usepackage{gensymb}
\usepackage{balance}

\usepackage{graphicx}
\usepackage{subcaption}

\usepackage{multirow}
\usepackage{cite}
\usepackage{amsmath,amssymb,amsfonts}
\usepackage{graphicx}
\usepackage{textcomp}
\usepackage{tabularx}
\usepackage{amsmath}
\usepackage{dblfloatfix}
\usepackage{siunitx}
\usepackage{booktabs}
\usepackage{longtable}
\usepackage{float}
\usepackage{gensymb}
\usepackage{balance}

\usepackage{graphicx}   
\usepackage{subcaption} 

\title{\LARGE \bf
Optimizing energy consumption for legged robot by adapting equilibrium position and stiffness of a parallel torsion spring

\author{Danil Belov$^{1}$, Artem Erkhov$^{1}$, Farit Khabibullin$^{2}$, Elisaveta Pestova$^{1}$, \\ Sergei Satsevich$^{1}$, Ilya Osokin$^{1}$, Pavel Osinenko$^{1}$ and Dzmitry Tsetserukou$^{1}$}

\thanks{$^{1}$The authors are with Intelligent Space Robotics Laboratory and Artificial Intelligence in Dynamic Action Laboratory, Skolkovo Institute of Science and Technology, Bolshoy Boulevard 30, bld. 1, 121205, Moscow, Russia
Email: {\tt\small \{Danil.Belov, Artem.Erkhov, Elisaveta.Pestova, Sergei.Satsevich, Ilya.Osokin, P.Osinenko, D.Tsetserukou\}@skoltech.ru\ }
  }

\thanks{$^{2}$Farit Khabibullin is with the Cognitive Modeling Center, Moscow Institute of Physics and Technology, Institutsky Lane 9, 141701, Dolgoprudny, Russia 
        {\tt\small khabibullin.fr@phystech.edu}}%
}

\begin{document}

\maketitle
\thispagestyle{empty}
\pagestyle{empty}

\begin{abstract}

This paper is dedicated to the development of a novel adaptive torsion spring mechanism for optimizing energy consumption in legged robots. By adjusting the equilibrium position and stiffness of the spring, the system improves energy efficiency during cyclic movements, such as walking and jumping. The adaptive compliance mechanism, consisting of a torsion spring combined with a worm gear driven by a servo actuator, compensates for motion-induced torque and reduces motor load. Simulation results demonstrate a significant reduction in power consumption, highlighting the effectiveness of this approach in enhancing robotic locomotion.

\end{abstract}


\section{INTRODUCTION}

In recent years, quadrupedal robots have shown remarkable advancements in traversing unstructured terrains and executing dynamic, acrobatic movements.
These capabilities are largely driven by the development of the outrunner motors with high torque density, that allow robots to perform complex maneuvers with precision.
However, one of the significant challenges associated with these motors is their limited energy efficiency due to the heat losses.
Given that a substantial portion of a quadrupedal robot's operation involves cyclic, stereotypical movements, optimizing energy efficiency in these scenarios is crucial for extending operational time and improving overall performance.

\begin{figure}[hbt]
    \centering
    \includegraphics[width=1\linewidth]{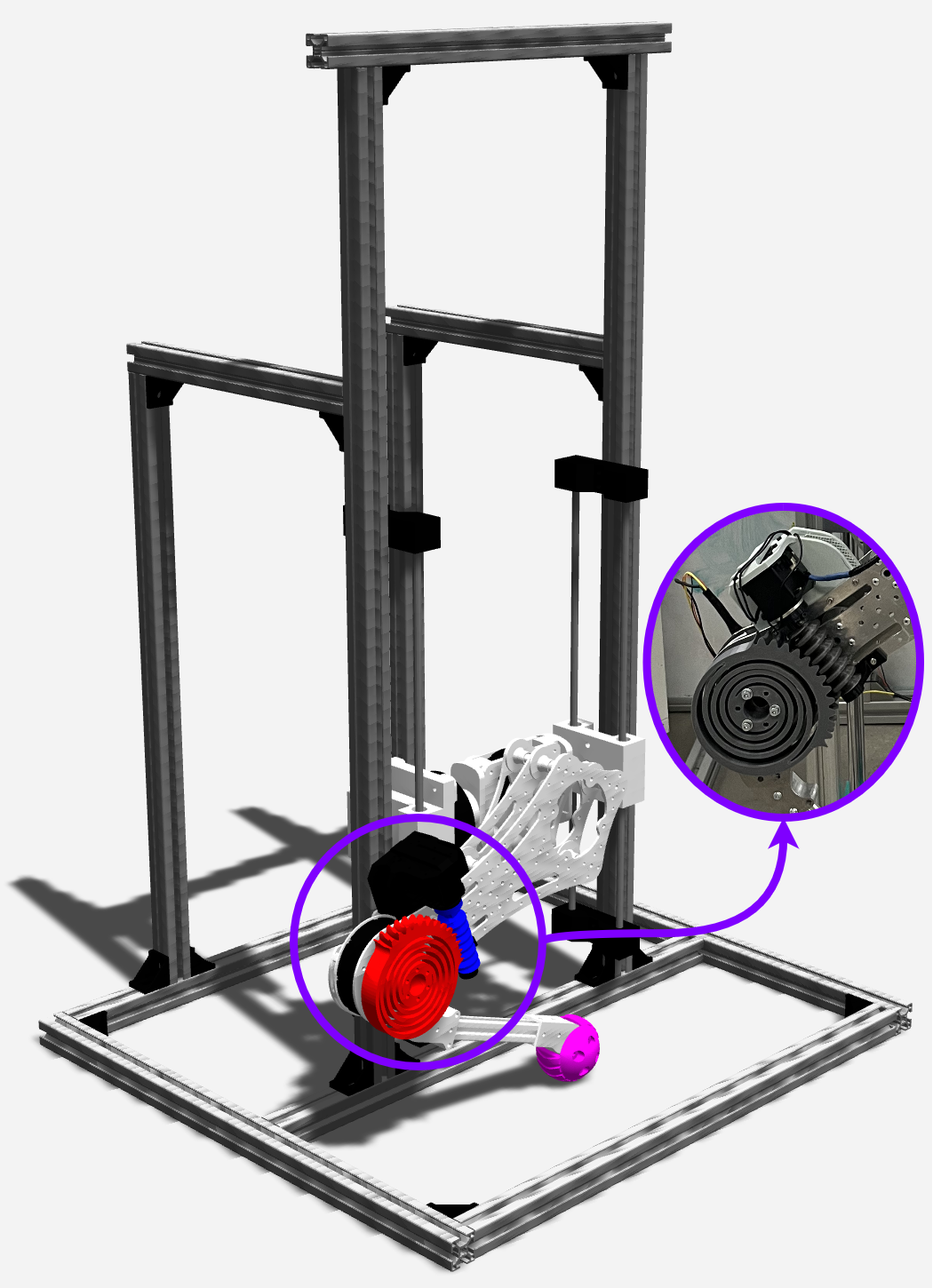}
    \caption{Digital twin of the one-leg stand with Adaptive Spring
System in Gazebo simulator.}
    \label{fig:sim.png}
\end{figure}

To address the energy efficiency challenge, parallel elastic elements have been widely incorporated into robotic joints, primarily for static load compensation and gravity compensation.
These elements reduce the energy required to maintain a particular position or to execute repetitive movements.
Part if the recent research is focused on enhancing the efficiency of the robots in the cyclic tasks by changing the compliance of these elastic elements.
While these solutions are promising, there remains a field for other types of adaptive mechanisms that can dynamically adjust to varying loads and environmental conditions.

In this work, an algorithm for compliance adaptation is proposed that utilizes a torsion spring with a variable equilibrium position.
This algorithm is designed to optimize energy consumption with the adjustment of a spring, which matches the real-time load conditions.

The contributions of this paper are as follows:

\begin{itemize}
    \item An algorithm for calculating optimal spring parameters is proposed. Specifically, the energy consumption functional minimization gives optimal stiffness and equilibrium position for a given cyclic task.
    \item A number of experiments in a simulator are carried out. The platform used is a 3DoF leg of a quadruped, that is attached to a stand, allowing vertical movements of the upper part of the leg (Fig. \ref{fig:sim.png}). In the experiments the energy consumption without and with an optimal spring are compared. Different loads, motion frequencies, motion amplitudes and starting positions are considered.
    \item It was thus shown that the proposed approach significantly reduces energy consumption in the simulated environment.
\end{itemize}

\section{RELATED WORK}

The development of adaptive compliance mechanisms in legged robots has been extensively studied, particularly focusing on enhancing energy efficiency during cyclic tasks.
Various approaches have been explored to integrate variable stiffness and parallel elastic elements into robotic joints in order to achieve this goal.

A key reference for the presented work is \cite{elastic4}.
Iterative design of a spring with RL in a loop was performed for a quadrupedal robot.
Starting with an initial guess on the spring parameters, an RL-based controller was trained.
After that, a set of trajectories was obtained.
Finally, optimization for finding optimal spring parameters was performed, and the process was repeated several times.
After convergence, a mechanical spring was produced and installed in the robot.

What if a spring could be adapted online, as the robot is working?
This is exactly the topic of this paper.

A significant contribution is a paper on nonlinear adaptive compliance, that aims at optimizing the energy consumption by adjusting the compliance in real-time according to task demands \cite{elastic1}.
This method relies on a relatively complicated assembly with 4 different springs, allowing for the adjustment of the spring stiffness.

Another approach involves the variable recruitment of parallel elastic elements, as discussed in the concept of Series-Parallel Elastic Actuators (SPEA) with dephased mutilated gears \cite{elastic2}.
This design enables selective engagement of elastic elements, allowing fine control over stiffness and improving energy efficiency across different operational conditions.

Among the works where deep analysis and extensive simulations are provided, \cite{elastic3} should be noted.
There the use of dynamic gravity cancellation and regulation control in robots with flexible transmissions is explored with an emphasis on employing variable stiffness to reduce energy required for maintaining postures and performing dynamic tasks.

Additionally, \cite{energy_efficient} provides an extensive examination of strategies for enhancing energy efficiency in robotic locomotion, emphasizing energy management in long-duration operations such as exploration and service tasks. 

Key strategies discussed in the paper include the implementation of compliant actuators, energy recovery mechanisms, and the optimization of gait patterns. They also highlight the value of biomimetic approaches, suggesting that inspiration from biological systems can lead to significant improvements in energy efficiency.

Adaptation in variable parallel compliance has also been a focus of research, where the equilibrium position of elastic elements is dynamically adjusted to minimize energy consumption during cyclic tasks \cite{elastic5}.
This method is particularly beneficial for tasks involving varying loads and stiffness requirements throughout the cycle.
It is important to note that the convergence and optimality of the presented method are proved in the paper.

Similarly, research on variable compliance in rotary mechanisms has demonstrated how adaptive control of elastic elements can lead to significant energy savings, especially in applications involving rotary motion \cite{elastic6}.

The interplay between compliance and task frequency is another area of study, with research showing that optimizing both stiffness and task frequency can lead to maximal energy efficiency in high-frequency cyclic movements \cite{elastic7}.
A comprehensive review of variable stiffness actuators provides an overview of different design approaches and components used to implement variable stiffness, highlighting their potential advantages and challenges \cite{elastic8}.

The work \cite{elastic99} introduces the Adjustable-Equilibrium Parallel Elastic Actuator (AE-PEA), a significant advancement over traditional parallel elastic actuators (PEAs). Unlike conventional PEAs, which have a fixed equilibrium position, the AE-PEA allows for real-time adjustment of the equilibrium position without requiring additional energy. This adaptability enhances its application in robotics, particularly in tasks requiring variable load positions. The design, which incorporates 3D-printed composite springs, offers customizable stiffness, making the AE-PEA more versatile and energy-efficient compared to existing actuators like Series Elastic Actuators (SEAs) and Series Variable Stiffness Actuators (SVSAs). This innovation represents a valuable contribution to actuator technology, with potential implications for improving energy efficiency in robotic systems.

Finally, in Minimum Gain Requirements for Trajectory Tracking of Compliant Robots in Divergent Force Fields the analysis of tracking performance with adaptive compliant elements is presented \cite{barat2023minimum}.
Additionally, simulation is performed for 1D and 2D systems.

While existing approaches have made significant strides in enhancing energy efficiency through the use of adaptive compliance and variable stiffness, the proposed approach offers a number of advantages.

The proposed system system relies on a torsion spring with an adjustable equilibrium position directly within the knee joint.
This enables the derivation of a closed-form solution for the optimal spring parameters, namely the equilibrium position and the stiffness.


Compared to the variable recruitment of parallel elastic elements as explored in \cite{elastic2}, the proposed approach is simpler, eliminating the need for complex engagement mechanisms.
Instead, a single adaptive torsion spring is used, providing compensatory torque, directly counteracting the torque from the load.
This approach reduces the mechanical complexity of the system and ensures a more consistent and predictable reduction in energy consumption.

While learning-based methods \cite{elastic4} have the potential to adaptively improve performance over time, they require extensive training data and computational resources.
Furthermore, this approach does not support real-time spring adaptation.

The article \cite{sharbafi2020parallel} discusses parallel elastics in legged robots to improve locomotion efficiency and robustness, highlighting how integrating elastic elements can reduce energy consumption and enhance stability.

This approach provides a theoretical foundation and practical applications for designing more effective and resilient robotic systems. Also work \cite{liu2024novel} examines the integration of parallel elastic actuators in robotic systems to enhance performance. It highlights how adding elastic elements in parallel with motors can reduce energy usage and improve the robot's adaptability to different terrains. The study provides both a theoretical framework and practical examples, demonstrating that parallel elastics can significantly boost the efficiency and robustness of legged locomotion.

In summary, the proposed method is straightforward, computationally lightweight and presumably relatively easy to implement in the hardware, see future work \ref{future_work}.


\section{PROPOSED APPROACH}

\subsection{System Overview} 

To validate the performance of adaptive mechanism, a simulation environment was designed to replicate a simplified experimental setup, that includes a robotic leg with three degrees of freedom, similar to the one that will be used in a real quadrupedal robot. The simulation setup consists of virtual stand, that holds the robotic leg and restricts its movement to a vertical axis, simulating the up-and-down motion without requiring the leg to maintain balance. This simplification focuses the study on the effectiveness of the torsion spring mechanism in the reduction of the energy consumption.

The 3D model of the stand and the robotic leg was initially assembled in Fusion 360. This model was then converted into a URDF (Unified Robot Description Format) file using a specialized script \cite{fusion2urdf}. This conversion was crucial for integrating the model into the Gazebo simulation environment, allowing for the accurate replication of the physical properties and dimensions of the leg. The lengths of the thigh and shin links of the leg is 0.28 meters each.

The torsion spring is modeled in the simulation using a Gazebo plugin specifically designed to replicate the behavior of a torsion spring in the knee joint. A plugin \cite{gazebo_plugin}, allowing to introduce a controlled torsional spring that could be adjusted during experiments, was used. This plugin simulates the effect of a real torsion spring by applying forces that mimic the spring's behavior based on its stiffness and equilibrium position.

The control of the robotic leg within the simulation is managed through ROS (Robot Operating System). Trajectories for the leg's movement are predefined in Python scripts encapsulated within ROS packages. These trajectories are translated into motor commands, which are then executed within the simulation environment. The feedback from the simulation, including joint angles and torques, was collected during the experiments.




\subsection{Optimal spring parameters derivation}

Let $\{(\alpha_i, \tau_i)\}^T_{i=1}$ be a T-step motion of the knee actuator, where $\alpha_i$ is the angle at the knee joint and $\tau_i$ is the torque exerted by the actuator at that moment.
It is assumed that number of full cyclic motions was performed.

With the uniform time steps, the total energy consumed by the actuator over this trajectory can be expressed in the form of Equation \ref{eq:1}, where $K$ is a motor-specific constant.

\begin{equation} \label{eq:1}
E = K \sum \limits_{i=1}^T \tau_i^2 \cdot \Delta t
\end{equation}

Introducing a linear torsion spring with an equilibrium position of $\alpha_0$ and stiffness $\mu$ changes the energy consumption.
With the proper parameters, the spring partially compensates the torque required from the actuator.
The energy consumption with a spring reads Equation \ref{eq:2}.

\begin{equation} \label{eq:2}
E = K \sum \limits_{i=1}^T (\tau_i - \mu (\alpha_i - \alpha_0))^2 \cdot \Delta t
\end{equation}

Here $\mu (\alpha_i - \alpha_0)$ is the torque produced by the spring, that reduces the torque that the actuator itself has to produce.

Let us differentiate the total energy consumption with respect to the spring parameters.

\begin{equation} \label{eq:3}
\left\{
\begin{array}{lr}
\dfrac{\partial E}{\partial \mu} = 2 K \sum \limits_{i=1}^T (\tau_i + \mu (\alpha_0 - \alpha_i)) (\alpha_0 - \alpha_i) \cdot \Delta t = 0\\
\dfrac{\partial E}{\partial \alpha_0} = 2 K \sum \limits_{i=1}^T (\tau_i + \mu (\alpha_0 - \alpha_i)) \mu \cdot \Delta t = 0
\end{array}
\right.
\end{equation}

\medskip

Solving this system for $\mu$ and $\alpha_0$ yields Equations \ref{eq:4}.

\medskip

\begin{equation} \label{eq:4}
\left\{
\begin{array}{lr}
\mu^* = \dfrac{\sum \limits_{i=1}^T \alpha_i \sum \limits_{i=1}^T \tau_i - N \sum \limits_{i=1}^T \alpha_i \tau_i}{\left(\sum \limits_{i=1}^T \alpha_i\right)^2 - N \sum \limits_{i=1}^T \alpha_i^2} \\
\alpha_0^* = \dfrac{\sum \limits_{i=1}^T \tau_i \sum \limits_{i=1}^T \alpha_i^2 - \sum \limits_{i=1}^T \alpha_i \tau_i \sum \limits_{i=1}^T \alpha_i}{\sum \limits_{i=1}^T \alpha_i \sum \limits_{i=1}^T \tau_i - N \sum \limits_{i=1}^T \alpha_i \tau_i}
\end{array}
\right.
\end{equation}



\section{EXPERIMENTS AND RESULTS}

\subsection{Experimental setup}

A number experiments were conducted.
Reference trajectories were generated for the robot base height, that evolved over time as $h(t) = h_0 + A \cdot \sin \left(\frac{t}{T}\right)$.
In the experiments mass $m$ varies, as well as the period $T$, amplitude $A$ and starting height $h_0$.
In all the experiments the actuators were controlled by position with the torques being generated by a PD-controller with $K_p = 300$ and $K_d = 1$.
The friction between the leg and the surface was set to be negligibly small in order not to disturb the motion of the robot.
The control sampling frequency was set to be 100 Hz. Data was gathered during 10 seconds, equivalent to 1000 simulation ticks.

\begin{figure}[h!]
    \centering
    \begin{minipage}[b]{0.48\linewidth}
        \centering
        \includegraphics[width=\linewidth]{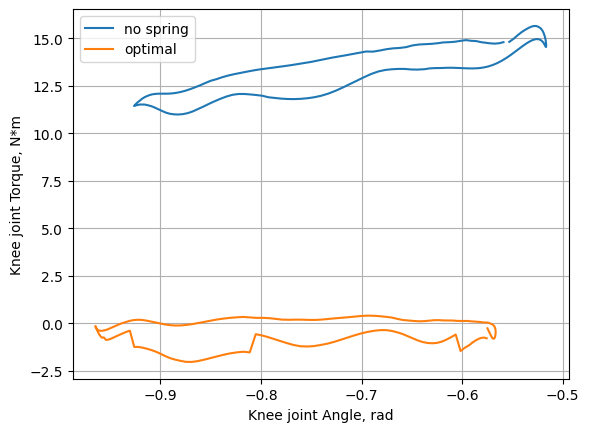}
        \caption*{Baseline \\  }
    \end{minipage}
    \hfill
    \begin{minipage}[b]{0.48\linewidth}
        \centering
        \includegraphics[width=\linewidth]{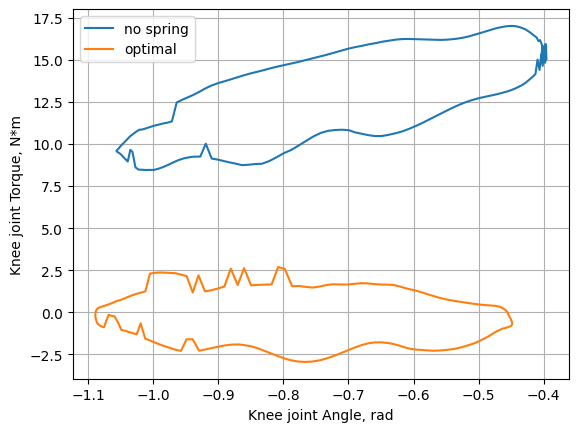}
        \caption*{Amplitude changed from 0.05 to 0.08}
    \end{minipage}

    \vspace{0.3cm} 

    \begin{minipage}[b]{0.48\linewidth}
        \centering
        \includegraphics[width=\linewidth]{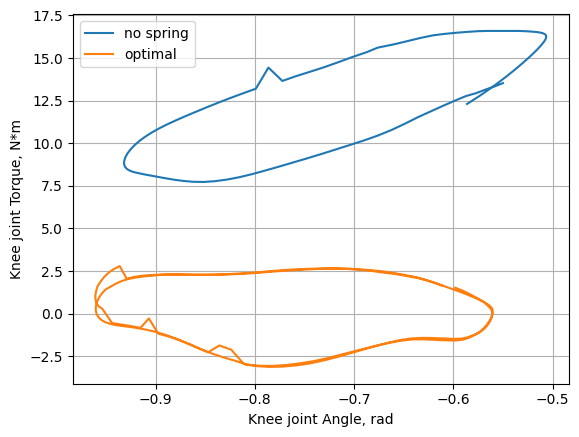}
        \caption*{Frequency increased from 100Hz up to 200Hz}
    \end{minipage}
    \hfill
    \begin{minipage}[b]{0.48\linewidth}
        \centering
        \includegraphics[width=\linewidth]{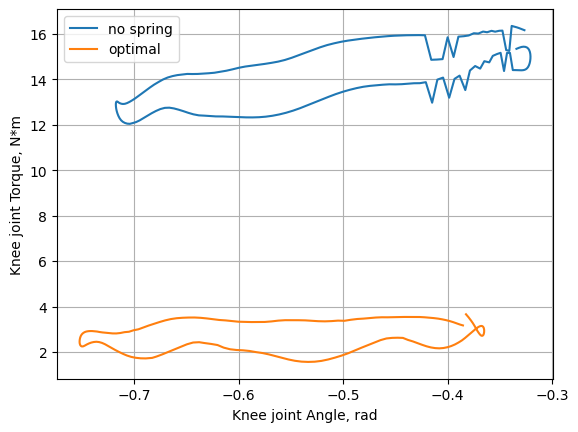}
        \caption*{Initial high changed from -0.2 to -0.15}
    \end{minipage}

    \vspace{0.3cm} 

    \begin{minipage}[b]{0.48\linewidth}
        \centering
        \includegraphics[width=\linewidth]{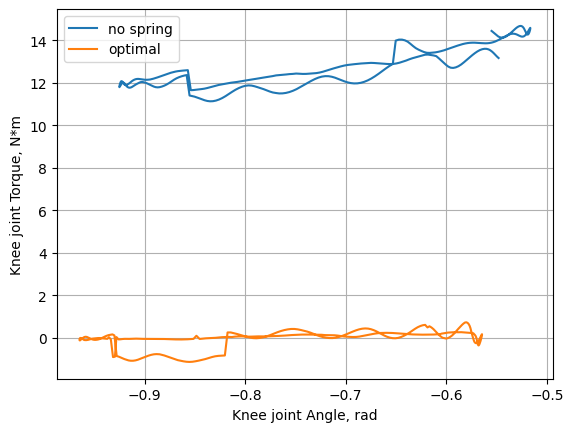}
        \caption*{Frequency decreased from 100Hz up to 50Hz}
    \end{minipage}
    \hfill
    \begin{minipage}[b]{0.48\linewidth}
        \centering
        \includegraphics[width=\linewidth]{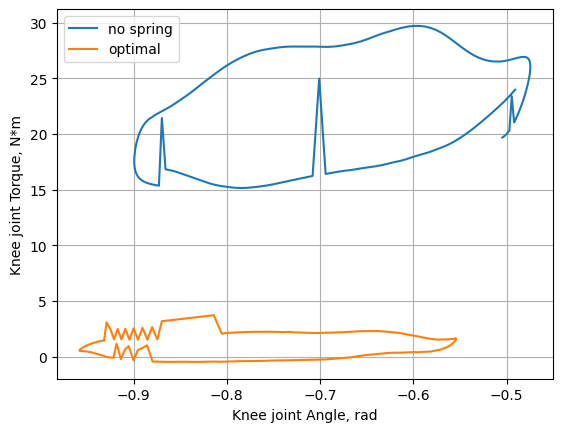}
        \caption*{Mass of the upper part increased up to 4 kg}
    \end{minipage}

    \caption{Comparison of knee motor torques during a single cyclic
    motion without and with a spring under different circumstances (varying mass, oscillation frequency, amplitude and starting height).}
    \label{fig:comparison}
\end{figure}

For each of the experiments without a spring the optimal $\mu^*$ and $\alpha_0^*$ were calculated.
After that, another simulation with a spring was performed.
The results are presented in the table \ref{table:results}.

\begin{table}[hbt]
\setlength{\tabcolsep}{5pt} 
\renewcommand{\arraystretch}{1.7}
\begin{tabular}{|c|c|c|c|c|c|c|c|c|} 
 \hline
 m & $T$ & A & $h_0$ & $E_0$ & $E_a$ & $\mu^*$ & $\alpha_0^*$ & $\frac{E_a}{E_0}$ \\ [0.5ex]
 \hline
 4.1 & 1.88 & 0.05 & 0.2 & 2285.9 & 14.2 & 8.54 & -2.23 & 0.7\% \\
 \hline
 4.1 & 1.88 & \bf{0.08} & 0.2 & 2114.2 & 30.2 & 8.77 & -2.15 & 1.4\% \\
 \hline
 4.1 & 1.88 & 0.05 & \bf{0.15} & 2627.1 & 103.1 & 7.05 & -2.52 & 3.9\% \\
 \hline
 \bf{8.1} & 1.88 & 0.05 & 0.2 & 6813.9 & 44.1 & 13.78 & -2.28 & 0.6\% \\
 \hline
 4.1 & \bf{0.94} & 0.05 & 0.2 & 2365.2 & 44.7 & 17.1 & -1.4 & 1.9\% \\
 \hline
 4.1 & \bf{3.77} & 0.05 & 0.2 & 4193.6 & 6.4 & 6.07 & -2.84 & 0.15\% \\
 \hline
\end{tabular}
\caption{Comparison of the robot's performance under different circumstances (varying mass, oscillation frequency, amplitude and starting height) with and without the spring that has optimal parameters $\mu^*$ and $\alpha_0^*$.}
\label{table:results}
\end{table}

\subsection{Results}

Figure \ref{fig:comparison} shows a comparison of the torques during a single cyclic motion with and without a spring under different conditions.
The spring parameters for the simulation were obtained from the system trajectories without a spring using a closed-form expression \ref{eq:4}.

\begin{figure}[h!]
    \centering
    \includegraphics[width=1\linewidth]{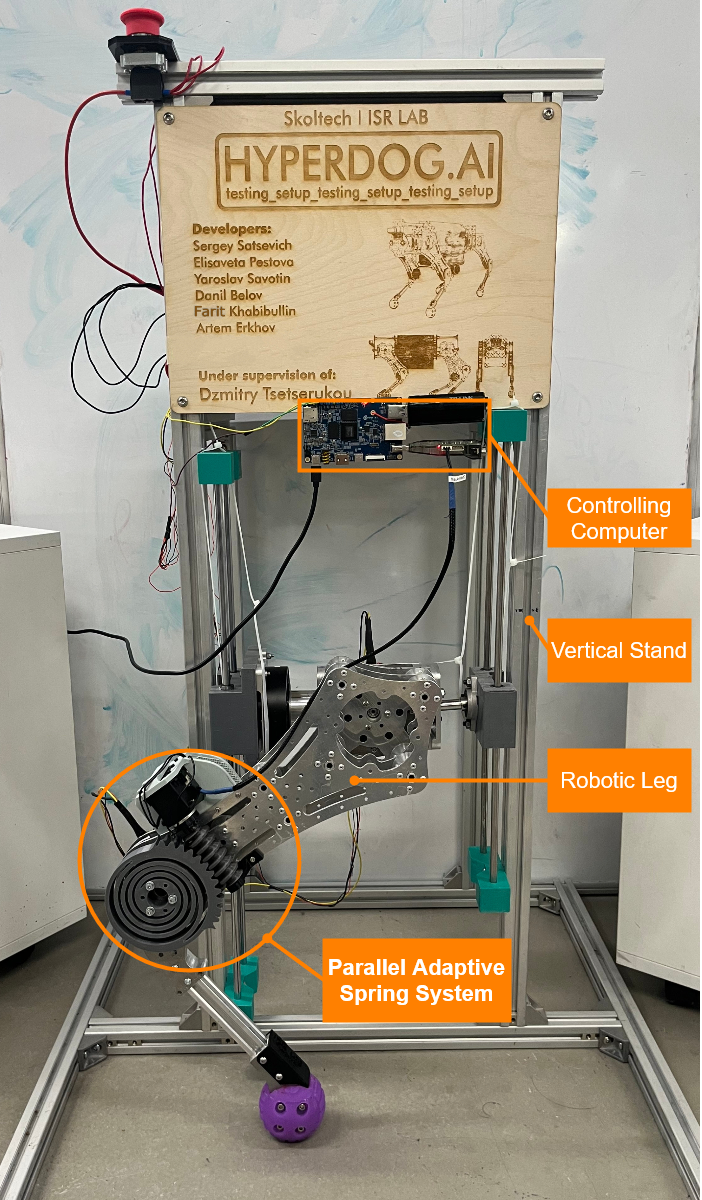}
    \caption{The experimental setup consists of a leg, fixing stand,
control electronics, a torsion spring mounted parallel to the
knee joint, and motor to control the preload on the spring.}
    \label{fig:real.png}
\end{figure}

\section{CONCLUSION}

It was thus shown that with an adaptive spring the overall energy consumption could be considerably reduced.
In the last column of the Table \ref{table:results} the ratio between optimized and  baseline energy consumption is given.
It is pretty clear that in the experiments with a physical stand the reduction in power consumption will be not as fascinating because the energy being lost due to friction.
However, it could be seen from the system trajectories that the overall load on the servos has decreased.
That not only means that less energy is used, but also leads to the reduction of wear of the physical servos.

\textbf {Code and other supplementary materials can be accessed by link: https://github.com/dancher00/AdaptiveSpring \cite{AdaptiveSpring}}

\section{FUTURE WORK}

\label{future_work}

\begin{figure}[h!]
    \centering
    \includegraphics[width=1\linewidth]{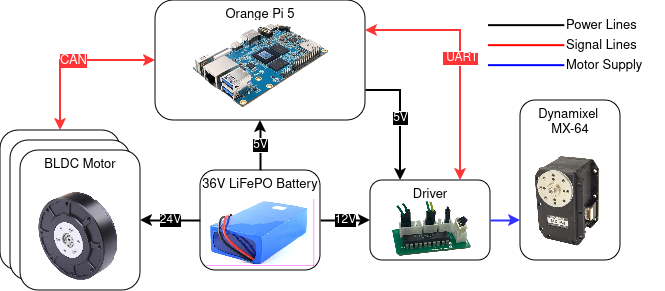}
    \caption{System overview, including modules and interfaces
developed for dynamical change of spring preload.}
    \label{fig:system.png}
\end{figure}

Future developments include the mechanical implementation of a proposed adaptive algorithm on a physical test bench. The vertical stand with robotic leg (Fig. \ref{fig:real.png}) with electronics (Fig. \ref{fig:system.png}) was well described and shown in the work \cite{HyperSurf}.

The core of adaptive system is the integration of a torsion spring attached to the knee joint of the robotic leg. This torsion spring is a critical component that allows for the adjustment of the leg’s compliance based on the load it experiences during operation. The spring's equilibrium position can be altered, providing an adaptive response to varying conditions and improving energy efficiency during cyclic movements.

The adjustment of the torsion spring is achieved through a Dynamixel servo motor, that controls the spring via a worm gear mechanism. As the servo rotates, the worm gear drives the rotation of the torsion spring, shifting its equilibrium position. That allows for the precise control of the compensatory torque exerted by the spring on the knee joint. Thus, the gravitational forces or external loads can be partially compensated, reducing the energy consumption of the motors during repetitive tasks.


\section*{Acknowledgements} 
Research reported in this publication was financially supported by the RSF grant No. 24-41-02039.

The authors would also like to thank Yaroslav Savotin for his great support in working on the wiring of the testing bench.

\addtolength{\textheight}{-12cm}



\end{document}